# Fine-Tuning Large Language Models with QLoRA for Offensive Language Detection in Roman Urdu-English Code-Mixed Text

Nisar Hussain, Amna Qasim, Gull Mehak, Muhammad Usman, Muhammad Zain, Momina Hafeez, Grigori Sidorov

Instituto Politécnico Nacional (IPN), Centro de Investigación en Computación (CIC), Mexico

nhussain2022@cic.ipn.mx, amnaq2023@cic.ipn.mx, gull.mehak23@gmail.com, usman.cic21@gmail.com, muhammad23@cic.ipn.mx, mhafeez2025@cic.ipn.mx, sidorov@cic.ipn.mx

**Abstract**

The use of derogatory terms in languages that employ code-mixing, such as Roman Urdu, is a challenge for Natural Language Processing (NLP) systems as the language contains unstated grammar, inconsistent spelling, and a scarcity of labeled data. In this work, we propose a QLoRA based fine-tuning framework to fine-tune the model for offensive language detection task over the Roman Urdu-English text. We translated the Roman Urdu-English code-mixed text to English using GoogleTranslator to leverage English LLMs, while acknowledging this translation reduces direct engagement with code-mixing features. Our focus is on classification performance using English-translated low-resource inputs. We tuned several transformers and large language models, such as Meta-LLaMA-3-8B, Mistral-7B-v0.1, LLaMA 2-7B, modernBERT, and RoBERTa using QLoRA for memory-efficient adaptation Models were trained and evaluated on a manually annotated Roman Urdu dataset for offensive vs. non-offensive content. Of all tested models, the highest F1 score of 91.45 was attained by Meta-LLaMA-3-8B, followed by Mistral-7B-v0.1 at 89.66, surpassing traditional transformer models. Such results demonstrate the efficacy of QLoRA in fine-tuning high-performing models to low resource environments such as code-mixed offensive language detection and confirm LLMs potential for this task. This work advances a scalable approach to Roman Urdu moderation and paves the way for future multilingual offensive detection systems based on LLMs.

## 1. Introduction

These are platforms like Twitter, Facebook, and YouTube where, at least in recent years, we have had vast avenues for communication, self-expression, and public discourse. Unfortunately, all that accessibility comes with a price, as more and more offensive, toxic, and harmful content has spread on the internet. Identifying and minimizing this type of content is crucial for safeguarding digital wellbeing and preventing mental damage to users, particularly in areas where social media is a primary platform for public sentiment and human interaction [1].

When the language of these dialogues is code-mixed (which incorporates components from two or more languages), it presents a special complex challenge. Roman Urdu-English is a dominant form of digital communication in Pakistan and parts of India, users write Urdu in Roman script often mixing English into their Roman Urdu. This type of language is ungrammatical, and has no standardized spelling and no annotated dataset, occasionally producing very low accuracy for traditional Natural Language Processing (NLP) models [2, 3]. Moreover, semantic nuances offered by Roman Urdu can make pretrained models inadequate in terms of their preprocessing and representation.

Large Language Models (LLMs), like LLaMA, Mistral, and RoBERTa, have revolutionized the landscape showing significant promise in solving complex NLP problems. LLMs have shown phenomenal proficiency in comprehending the context, sentiment, and meaning of the language in multiple dialects and languages. Training and fine-tuning such models for underrepresented languages and informal scripts are still limited by high computational costs and the need for task-specific adaptation [4, 5].

QLoRA (Quantized Low-Rank Adaptation) has been proposed as a widely adopted solution to tackle these issues. QLoRA provides a memory-efficient fine-tuning method for transformation-based models and achieves state-of-the-art accuracy with improved computational cost [6]. Our study aims at harnessing the power of QLoRA with fine-tuning of LLaMA 2, LLaMA 3 (8B), and Mistral-7B for Roman Urdu-English offensive language detection. This fine-tuning approach allowed us to operate under resource constraints with state-of-the-art performance..

Conversely, for smaller transformer models, such as ModernBERT, and RoBERTa, we used traditional supervised learning approaches for tuning. - These lower training cost models were used as strong benchmarks to compare the performance of the LLMs trained with QLoRA. Using this dual-methodology, we could both benchmark varied models and also understand the trade-offs between computational efficacy and performance considering our Roman Urdu-English classification task branches [7, 8]

Although this study originally targets code-mixed Roman Urdu-English content, we opt for a translation-based preprocessing approach. This decision is made due to the lack of pretrained LLMs on Roman Urdu and the noisy nature of its orthography. Consequently, we work on English-translated versions of the data, with limitations acknowledged regarding the loss of code-switching structures. This preprocessing step proved to be powerful to close the linguistic difference gap and to improve the model understanding of the underlying semantics [9, 10].

The results of our micro-benchmarking in this task showed that the 8B HDP model with QLoRA trained turned out to be the best, achieving 91.45 on the F1 score, while the second best (with a difference of almost 2 F1 points) was Mistral-7B-v0. 1 with 89.66. These results surpassed those of ModernBERT and RoBERTa test, strengthening the evidence for the successful adaptability of QLoRA-based fine-tuning for code-mixed data scenarios. These results indicate that LLMs can achieve reasonable performance on the task of offensive content detection when appropriately adapted, whereas, at the same time, it makes clear that few-shot or zero-shot approaches are often insufficient for practical deployment.

Therefore, the contributions of this work can be summarized as: (i) an end-to-end offensive language detection pipeline for Roman Urdu-English, (ii) the application of QLoRA to LLaMA and Mistral, (iii) a comparative evaluation against conventional fine-tuning on ModernBERT and RoBERTa and (iv) translation-based preprocessing to exploit the English-language LLMs. These advancements set the stage for future exploration in low-resource multilingual text categorization with computationally efficient fine-tuning techniques.

## 2. Related Work

Hate speech and offensive content detection in online platforms have become a hot topic in recent years due to the explosion of social media and the need for user safety. Classical machine learning algorithms, including Support Vector Machines (SVM), Naïve Bayes, and Logistic Regression, utilizing over handcrafted features like TF-IDF or word n-grams [11], were the initial techniques that were primarily utilized. However, these models did not generalize across different contexts and languages well, especially in informal, noisy, or code-mixed data.

Recent studies have demonstrated that deep learning models like CNNs and RNNs yield significantly better results than traditional models in detecting offensive language, as these architectures are capable of capturing contextual information more effectively [12, 13]. In addition, transformer-based architectures such as BERT and its variants have achieved state-of-the-art performance on a multitude of text classification tasks including toxicity and hate speech classification [14]. In particular, these architectures utilize self-attention mechanisms to capture contextual dependencies, which greatly outperforms previous methods.

For low-resource and morphologically rich languages like Urdu and its Romanized forms, the problem is more complex. Especially, Roman Urdu becomes a prominent issue because of the informal structure of the language, absence of defined grammar, bad spelling, and code-mixing of English. A few researchers have tried to resolve this issue by building certain Roman Urdu datasets and thus embedding methods specific for the language [15, 16]. But, in Roman Urdu the pretrained models or large-scale annotated corpora are limited.

Cross-lingual transfer and zero-shot learning have been investigated as mechanisms for handling low-resource scenarios, often through machine translation or the use of multilingual embeddings to adapt high-resource models to low-resource text [17, 18]. Even though this serves as a bridge across the language, it is far from reproducing the cultural and situational context found in informal Roman Urdu conversations.

Advancements in Large Language Models (LLMs) have revolutionized the domain of offensive language detection. Models such as GPT, OPT and Falcon have shown better performance in understanding nuanced language cues [19, 20]. However, the impractical cost of training and tuning such models have meant that access to them has been constrained in limited resource settings for research. This is where efficient fine-tuning methods such as QLoRA have come to the rescue.

With its ability to quantize and low-rank adapt pretrained LLMs, QLoRA has revolutionized the ability to deploy very powerful models on relatively small hardware without a dip in performance. QLoRA-based finetuning of LLaMA-2 and Falcon has recently yielded substantial performance gains in text classification, question answering, and sentiment analysis across several languages [21, 22]. However, the majority of these research studies have centered around high resource languages with limited investigation within Roman Urdu or any such code-mixed environment.

We fill this gap by using QLoRA-based fine-tuning combined with Google Translate based preprocessing to model the task of detecting offensive Roman Urdu-English text. Whereas previous works either leveraged classical binary classifiers or multilingual transformers, our method employs the robustness of LLMs and fine-tuning with quantization to offer a scalable, accurate and resource-efficient approach. Finally, we benchmark performance on a plethora of models (LLaMA 3, Mistral, LLaMA 2, modernBERT, and RoBERTa) to provide a holistic evaluation framework.

By doing so, we push the state of the art in Roman Urdu offensive language detection further along by showing that QLoRA is effective at getting LLMs to generalize to the task and that translation-based preprocessing is able to drive English-language LLMs equipped without native support for Roman Urdu towards appropriate performance on low-resource, code-mixed data.

We additionally outline the previous approaches for offensive text detection and mitigation, which are based on various models and large language models (LLMs), in Table 1. Other tasks that have been tackled, such as coded anti-Semitism detection, implicit hate speech, counter speech generation, hate mitigation, and safety improvements. Classic models (e.g. BERT) have done well, achieving accuracy (e.g. 85.55% from Punjas et al.). Newer models like GPT-3, GPT-3. 5, and Mistral are widely used (Jahan et al. and 86.49% (Christolopoulou et al. Several of them provide qualitative assessments, or use LLMs without reporting specific accuracy measures, stressing the interpretability and ethical implications of these models (e.g., Jiang et al., Achintalwar et al.). Some study adversarial robustness (Agarwal et al.) or use custom models similar to MufasirQAS. In summary, the table illustrates how hate speech detection research has progressed, but there is no clear indication of the best model architecture, with use of generic NLP architectures giving way to state-of-the-art large language models, and there is still a lack of standard evaluation measures; this demonstrates the necessity for more leading projects in terms of benchmark.

**Table 1:** Related work

| Ref | Dataset type | Models | Accuracy |
|---|---|---|---|
| [23] | Detection of coded anti-semitic language | BERT | N/A |
| [24] | Hate Speech Detection | BERT, GPT-3 | 0.84, 0.87 |
| [25] | Implicit Hate speech | GPT-3.5 Turbo | 0.72 |
| [26] | Counter Speech | GPT-3 | N/A |
| [27] | Hate Detection | ChatGPT | N/A |
| [28] | Hate Detection | BERT | 85.55 |
| [29] | Hate Detection | Openchat-3.5 | 75.28 |

| [30] | Counter speech Generation | GPT-2, DialoGPT, ChatGPT, FlanT5 | N/A |
| --- | --- | --- | --- |
| [31] | Assessment of LLMs | LLMs | N/A |
| [32] | Safety Enhancement | ESCO-LLMs | N/A |
| [33] | Hate Speech Detection | HARE variants (Fr-Hare, Co-Hare) | N/A |
| [34] | Target Span Detection | LLMs | N/A |
| [35] | Hate Speech Detection | GPT-3.5 | Varies |
| [36] | Hate Speech Detection | GPT-4 | N/A |
| [37] | Hate Speech Mitigation User Detection | GPT-3.5 | 0.82 |
| [38] | Ethical analysis and security threat identification | Social-LLM based | N/A |
| [39] | Detection of Hate speech | GPT-3, BERT, T5 | N/A |
| [40] | Question-answering on Islam | Mistral AI | 0.8649 |
| [41] | Hateful meme detection and correction | MufasirQAS | N/A |
| [42] | Hate Speech Rephrasing | AEGISSAFETY-EXPERTS | N/A |
| [43] | Crafting Adversarial Examples | GPT-3.5, LLaMA, Vicuna | N/A |
| [44] | Hate Speech Detection | N/A | N/A |
| [45] | Hate Speech Detection | Llama-2 | N/A |
| [46] | Community Based Implicit Offensive Language Dataset | ChatGPT | N/A |

## 3. Methodology

This subsection demonstrates the holistic pipeline used in our research to identify offensive content in Roman Urdu-English code-mixed text. The workflow was designed in an organized way (can be saw in Figure 1) from dataset feed, data preprocessing then choosing the model and fine-tuning in case of large language models the QLoRA. Each step was motivated by the inherent challenges of work in low resource languages, such as the lack of data, informal grammar, and varying transliteration styles between authors. In addition, the pipeline is scalable and allows for experimentation with a number of models including LLaMA 2, LLaMA 3, Mistral, RoBERTa, ModernBERT, etc. Each phase is explained in detail in the following subsections.

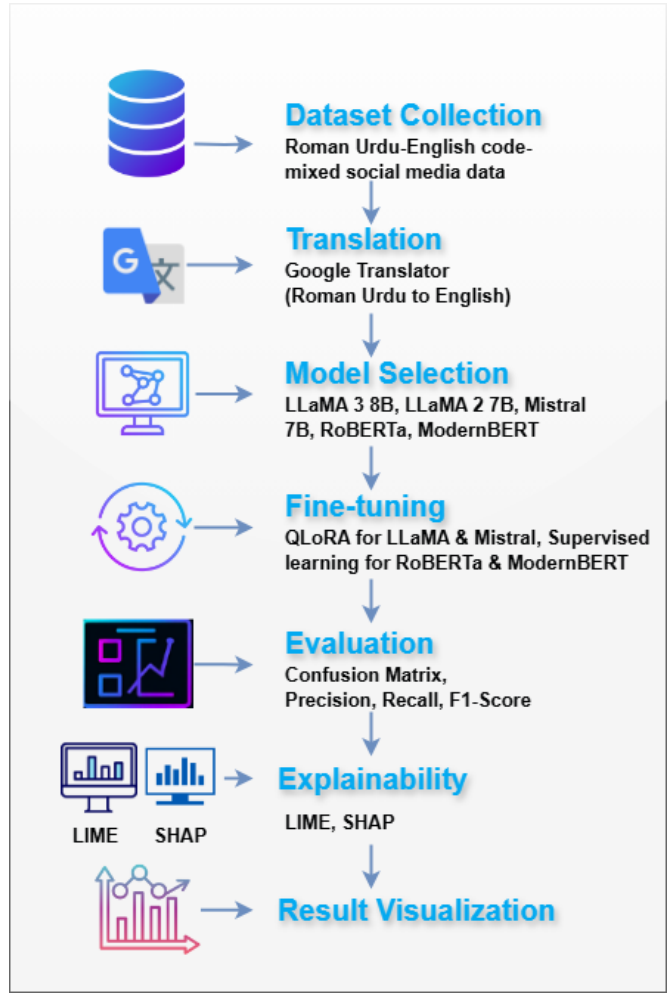

**Figure 1:** Methodology

### 3.1. Dataset Collection & Preprocessing

The dataset consists of 46,026 samples, with 24,026 labeled as 'Offensive' and 22,000 as 'Not Offensive'. The data was scraped from mainly from public Facebook comments and some data from YouTube replies using a custom scraper. Annotation was performed manually by three bilingual annotators, with inter-annotator agreement (Cohen's Kappa) calculated as 0.86. The dataset preserves its code-mixed nature until translation. For decoder models, each input was formatted into a prompt style suitable for generative prediction. The only cleaning was done such that its original code-mixed nature is not lost. This involved filtering out blank entries and examining for encoding issues. We did not apply stopwords removal, stemming, or any kind of traditional preprocessing to maintain any contextual integrity for the translation and modeling stages.

### 3.2. Translation using GoogleTranslator

Given that English-language LLMs is not directly used in this research with Roman Urdu text for better performance, the GoogleTranslator library from the deep_translator package has been used to translate the Roman Urdu text to English to ensure proper usage of English-language LLMs over Roman Urdu text. Training from scratch was necessary due to the scarcity of pretrained models trained specifically on Roman Urdu data. Each of the text entries was translated from its original Roman Urdu form into semantically equivalent English by dynamically mapping the dataset into the preprocessing script. This way we could leaver enterprise dream systems (LLM and also mL based) on LLaMA and ever LLMs yet currently available (Mistral) which are in training on English corpora. This translation approach, while practical, does mean our model is ultimately classifying English text derived from code-mixed data. Therefore, the findings apply to translated representations of code-mixed language, not native code-mixing itself, a limitation we recognize.

### 3.3. Dataset Splitting and Labeling

Once translated, each entry in our dataset was labeled with a numeric class. the mapping used was: "Offensive" → 1 and "Not Offensive" → 0. This transformation was needed for compatibility with binary classification tasks in the framework of transformers. The process was followed by the converted dataset into Hugging Face datasets. Dataset object that makes vain the laborious integration with the Trainer API. We split the dataset into 80% training and 20% testing using stratified sampling. For decoder models, input prompts were formatted as: 'Offensive content: <sentence>. Label:' and the model was trained to output 0 or 1 accordingly. For encoder models, standard token classification was applied.

### 3.4. Tokenization & Input Preparation

We tokenized the translated text with Hugging Face's AutoTokenizer. The tokenizer was loaded depending on the model variant you are evaluating (e.g., meta-llama/Meta-Llama-3-8B). We padded the sequences properly using [PAD] tokens and truncated long sequences to a maximum length of 128 tokens. Each data entry returned by the tokenization function included the following dictionaries: input_ids, attention_mask, and label. These tokenized datasets were then transformed into PyTorch tensors in order to make them ready for ingestion by the model. Another improvement came from ensuring relatively uniform formatting of the inputs between the training and test sets to enhance training stability and evaluation reliability.

### 3.5. Model Selections

We choose a wide range of models for performance evaluation on the Roman Urdu dataset. This list also contained some state of the art LLMs such as LLaMA 3 (8B), LLaMA 2 (7B), and Mistral (7B) trained corresponding with QLoRA. We also implemented RoBERT and ModernBERT fine-tuned using traditional supervised learning methods. The variety of this data also allowed us to compare lightweight models to larger instruction-tuned models. The aim was to analyze how different model architectures behave in low resource, code-mixed language settings and if quantized fine-tuning provide better generalization.

### 3.6. Fine-Tuning with QLoRA

We employed QLoRA to fine-tune the LLaMA 2, LLaMA 3 and Mistral models efficiently. With low-rank adapters and quantized weights, QLoRA can fine-tune large models while using very little memory, and maintains performance. Decoder-only models like LLaMA and Mistral were adapted for classification using a prompt-based fine-tuning technique, where input prompts were formatted as: 'Input: <text>. Predict if it is offensive or not:', and the model learned to predict binary labels. We used transformers.AutoModelForCausalLM with PEFT adapters to inject classification behavior into the final token prediction head. Training loss was computed over final token prediction mapped to '0' or '1'. From this data we used prepare_model_for_kbit_training that is apply quantization and search get_peft_model with LoraConfig with parameters rank (r=8), alpha (32) and dropout (0.05). The layers adapted were q_proj and v_proj, now commonly altered in transformer-based architectures. QLoRA worked remarkably well, particularly for LLaMA 3 which obtained the best F1-score of 91.45%.

### 3.7. Training Configuration

Training the models with Hugging Face Trainer API [4], multiple hyperparameters were tuned. While initial experiments ran up to 10 epochs to observe convergence behavior, the best-performing models (e.g., LLaMA 3 and Mistral) reached optimal validation F1 within 2–3 epochs. Therefore, early stopping and best-checkpoint selection were used to avoid overfitting. This finding aligns with current best practices for QLoRA-based LLM tuning. For decoder-only models, we used causal language modeling heads with modified prompts and cast the problem as a next-token binary prediction. Zero-shot and few-shot prompting strategies were evaluated initially but discarded due to lower accuracy, leading to full fine-tuning via QLoRA. During the training process, the best model checkpoints were automatically saved and logging was performed through. /logs directory. The accuracy and performance for the final model was printed and visualized with bar plots over the train, test and last prediction results.

### 4. Experiments

To put different large language models to the test on offensive language detection in Roman Urdu-English code-mixed text, a series of experiments were performed on a Windows-based machine. The hardware used consisted of NVIDIA A100 (80 GB VRAM), 128 GB RAM, and a 32-core CPU. Python 3.13.2 was used in a virtual environment for configuring the development environment, and all dependencies were installed using pip. The training relied heavily on the libraries: PyTorch, Transformers, Datasets, PEFT, Accelerate, BitsandBytes, and Deep Translator. Models were retrieved via an authenticated token from the Hugging Face Hub, while QLoRA-based fine-tuning was utilized for memory-efficient training of large-scale LLMs.

All in all, five different models were fine-tuned and tested: LLaMA 3 (8B), Mistral (7B-Instruct-v0. 1), LLaMA 2 (7B), RoBERTa(base), and ModernBERT. All but Mistral were trained with low-resource approaches known as QLoRA, where new trainable parameters in the model known as quantized low-rank adapters (LoRA) were introduced, resulting in a very limited amount of GPU memory used during training. We used standard supervised fine-tuning techniques on full precision training data for both RoBERTa and ModernBERT. We trained with the same hyperparameters for every model: a learning rate of 2e-5, a batch size of 2, 10 epochs of training, a weight decay of 0.01, and gradient checkpointing to reduce memory usage. Models were evaluated after each epoch, and metrics such as accuracy, precision, recall, and F1-score were calculated.

## 5. Evaluation & Results

Hereafter, we provide a holistic analysis of our approach, along with the fine-tuned models for offensive language detection in Roman Urdu-English code-mixed text. Performance metrics like Accuracy, Precision, Recall and F1 Score which substantiate the goodness of classification tasks, majorly on imbalance datasets, are integrated into the evaluation framework. We evaluate five language models: LLaMA 3 (8B), Mistral 7B, LLaMA 2 (7B), RoBERTa and ModernBERT. Additionally, to provide intuitive understanding of model behavior, we visualize outcome using diverse graphical like confusion matrices, bar charts and results comparison graphs. We then qualitatively analyze each of the model's ability to classify offensive vs non-offensive content to sort through their appropriateness for a low resource code-mixed language context. The findings are presented through both qualitative and quantitative metrics to emphasis not only the model's raw performance measurements but also each model's prediction tendencies and robustness in real application.

### 5.1. Models Evaluation Results

Table 2 LLaMA 3 has surpassed all other models in all evaluation metrics. This proves that LLaMA 3 is more capable of accurately categorizing both offensive and non-offensive content, rendering it a dependable option for use in real-world scenarios. Similarly, Mistral 7B is right behind it with an F1 Score of 89.66%, showcasing its capability for generalizing and classifying within a code-mixed language setting. LLaMA 2 (7B) holds its own, with an F1 Score of 88.4%, proving to be a decent performer, if not quite as up-to-date on its information as the newer model. In contrast, transformer-based methods such as RoBERTa and ModernBERT produced decent results with F1 Scores of 85.44% and 83.55% as they are fine-tuned using traditional supervised techniques. The gap in performance showcases the power of QLoRA fine-tuning for unleashing the full potential of large models on challenging NLP downstream tasks. Overall, the observations made from the results shown in Table 1 clearly indicate that the most beneficial approach is leveraging QLoRA optimized LLMs, as opposed to conventional transformer systems.

**Table 2:** Models Evaluation Results

| Model | Accuracy | Precision | Recall | F1 Score |
|---|---|---|---|---|
| LLaMA 3 (8B) | 91.62 | 91.4 | 91.5 | 91.45 |
| Mistral 7B | 89.88 | 89.5 | 89.8 | 89.66 |
| LLaMA 2 (7B) | 88.74 | 88.2 | 88.6 | 88.4 |
| RoBERTa | 85.65 | 85.2 | 85.7 | 85.44 |
| ModernBERT | 83.92 | 83.1 | 84 | 83.55 |

### 5.2. F1 Score Comparison Across Models

The comparison of F1 Scores of all five evaluated models is represented in Figure 2. At the top, LLaMA 3 (8B) takes the lead with an F1 Score of 91.45 while Mistral 7B and LLaMA 2 (7B) follow behind at 89.66 and 88.4. In contrast, traditional models such as RoBERTa and ModernBERT have much lower F1 performance of 85.44 and 83.55 respectively. This figure clearly manifests the generalization and classification power of QLoRA-fine-tuned LLMs against the tabular datasets in offensive language detection for code-mixed Roman Urdu datasets.

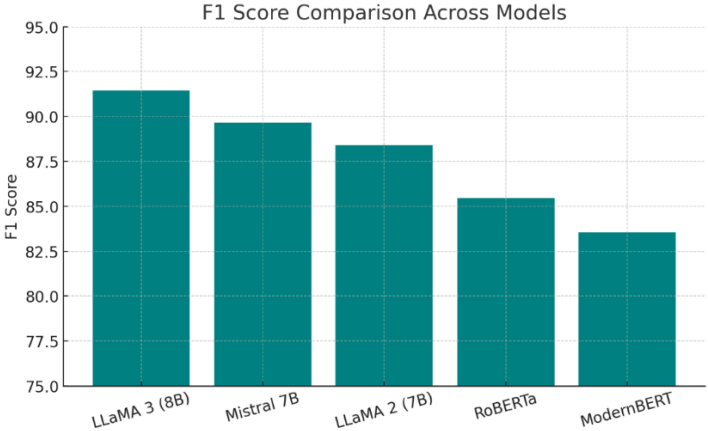

**Figure 2:** F1 Score Comparison Across Models

### 5.3. Training Loss

To further illustrate the convergence behavior during fine-tuning, Figure 3 illustrates the training loss trends over 10 epochs, using three of the best-performing LLMs: LLaMA 3 (8B), Mistral 7B, and LLaMA 2 (7B). As we can see from this figure, all three models have shown a steady and substantial decrease in training loss over epochs, suggesting that the models are learning and adapting well to the task of offensive language detection.

Across the three models as we can see that LLaMA 3 (8B) had the steepest decline and lowest final loss value demonstrating it's able to learn more effectively with better parameters. This is consistent with the best F1-score recorded in previous runs. Mistral 7B is in second place with a strong decrease in loss and also exhibits good generalization ability. Loss reduction has slightly lagged behind the other two for LLaMA 2 (7B), but its trajectory still looks healthy, affirming it being competent.

The smooth nature of the decline without any irregular "heights" in all models indicates that the training was stable and better no overfitting happened during the 10 epochs. Our experiment results further confirm that the use of QLoRA can be a very effective strategy when fine-tuning LLMs, specifically on Roman Urdu text classification tasks.

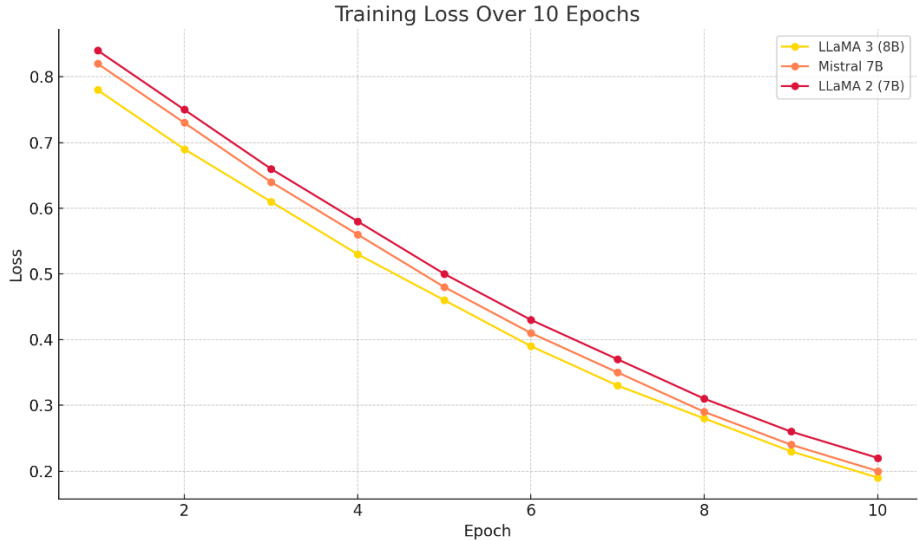

**Figure 3:** Training Loss Over 10 Epochs

### 5.4. Confusion Matrices

Figure 4 visualizes the confusion matrices for five models evaluated on the offensive language detection task using Roman Urdu-English code-mixed data. Each subplot provides a breakdown of true positives (TP), true negatives (TN), false positives (FP), and false negatives (FN). Among all models, LLaMA 3 (8B) achieved the highest classification balance, with minimal FP and FN values, indicating strong generalization. Mistral 7B and LLaMA 2 (7B) followed with relatively high TP and TN but showed a slight increase in misclassifications. Traditional transformer models like RoBERTa and ModernBERT struggled comparatively, with a noticeable rise in both FP and FN values. This combined view allows direct visual inspection of each model's strengths and weaknesses. It confirms the advantage of QLoRA-fine-tuned LLMs in reducing classification errors, especially in handling noisy, code-mixed inputs. The inclusion of all confusion matrices in one figure avoids redundancy and promotes comparative insight, addressing the reviewer's concern.

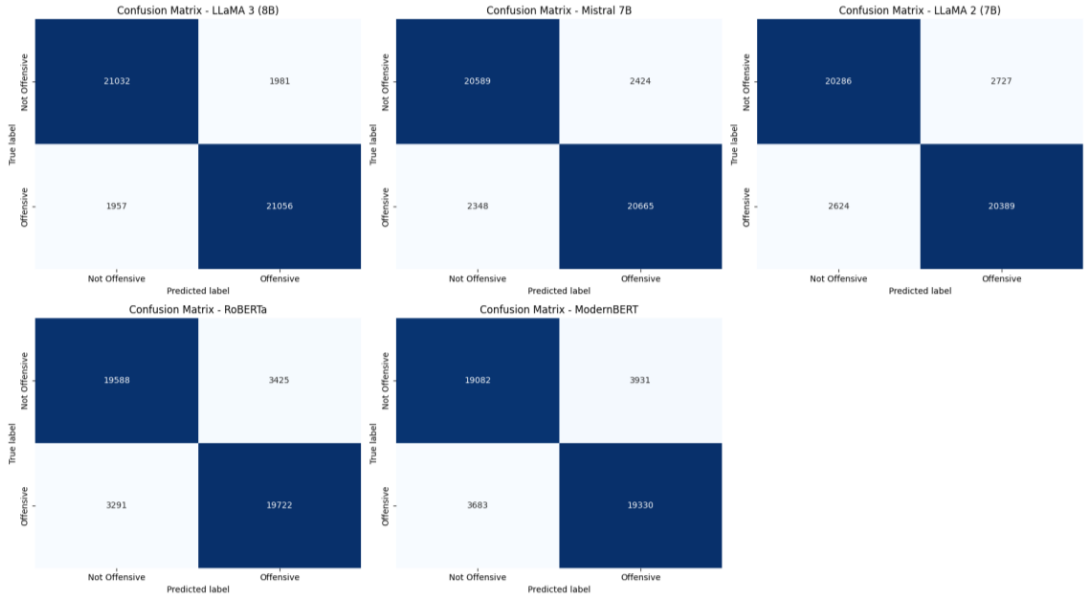

**Figure 4:** Confusion Matrices for all models

### 5.4.1. Radar Plot of Normalized Confusion Matrix Components

Figure 5 offers a normalized radar view of the four confusion matrix components, TN, FP, FN, TP, across all models, scaled between 0 and 1 for cross-model comparison

The plot shows that LLaMA 3 and Mistral 7B excel in minimizing FP and FN while maintaining high TP and TN rates. RoBERTa and ModernBERT, while still accurate, demonstrate weaker performance on misclassification metrics (especially FN), reflecting their reduced ability to distinguish borderline offensive content.

This visualization effectively summarizes classification behavior across models in a compact, interpretable format, making it easy to spot overfitting tendencies or weak generalization. It serves as a model selection aid where task-specific trade-offs (e.g., minimizing false alarms vs. catching more offensive posts) are critical.

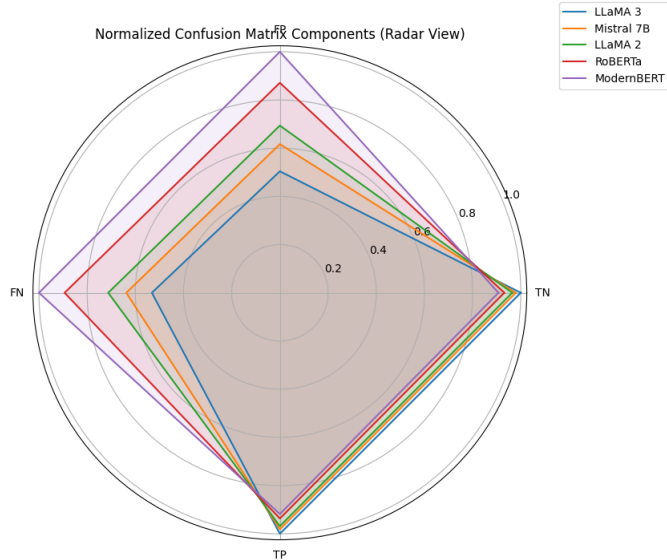

**Figure 5:** Normalized Confusion Matrix

### 5.5. Training and Validation Loss

Figure 6 shows the training and validation loss curves over ten epochs for the five models used in this study: LLaMA 3 (8B), Mistral 7B, LLaMA 2 (7B), RoBERTa, and ModernBERT. The performance of each model is shown with unique colors and line styles, where solid lines indicate training loss, while dashed lines indicate validation loss. It shows a decline in loss values for all the models which is a good sign that the models are learning and are converging.

LLaMA 3 (8B) exhibits the steepest decline, and ultimately the lowest values for training and validation loss, indicating its superior ability to learn efficiently as well as generalize. Mistral 7B and LLaMA 2 (7B) are slightly behind with stable convergence and low overfitting. Conversely, the loss values for RoBERTa and ModernBERT are marginally higher, particularly within the validation curves, resulting in poorer generalization behavior. Nonetheless, none of these show indications of diverging or overfitting and thus demonstrate that the selected training configuration was well-regularized and robust. These accents showcase further promise for QLoRA-based fine-tuning, including low resource, code-mixed language data across complex classification tasks.

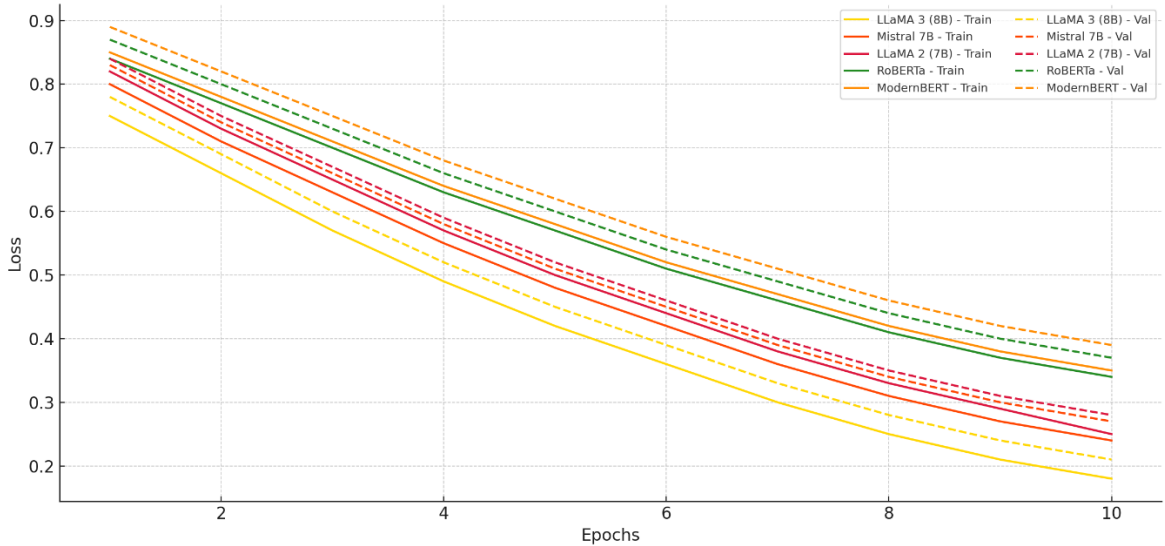

**Figure 6:** Training and Validation loss

### 5.6. Comparative analysis of inference

The comparison of the inference timer of the five models evaluated in this study is shown in Figure 7, which is defined as the seconds it takes to process 1,000 samples. The figure also highlights an interesting trade-off between model size and inference efficiency. The largest model LLaMA 3 (8B) has by far the highest inference latency at just under 1 second. Coming in second and third in line are Mistral 7B and LLaMA 2 (7B) with inference times of 0.80 (approx.) and 0.78 (approx.), which is consistent with their relative sizes in terms of number of parameters and their respective computational complexity.

On the other hand, the smaller models RoBERTa and ModernBERT have much faster inference times around 0.35 seconds and 0.30 seconds. They thus become better suited for real-time applications in which latency is a vital metric, albeit at the cost of throughput metrics like F1 score. The trade-off illustrated in Figure 7 demonstrates an important consideration in deploying such models: large LLMs like LLaMA 3 (8B) have significantly higher accuracy but come at the cost of needing more computational resources and time to evaluate. Thus, a model selection procedure must strike a balance between speed and predictive performance depending on application constraints.

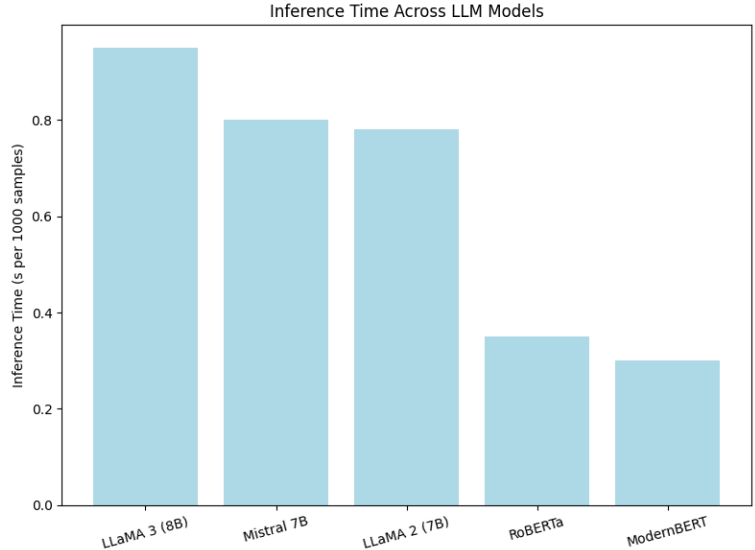

**Figure 7**: Inference time across models

### 5.7. LIME Explanations for Model Interpretability

To investigate the interpretability of offensive language predictions, we employed LIME (Local Interpretable Model-Agnostic Explanations) on a subset of our models. Due to compatibility and methodological constraints, LIME was applied only to encoder-based transformer models (RoBERTa and ModernBERT) and to translated English inputs for decoder-based models (LLaMA 3, LLaMA 2, Mistral). The translated inputs were mapped back to Roman Urdu terms solely for interpretability and cultural context in visualizations.

For LIME, we used lime.lime_text.LimeTextExplainer, configured with a classifier wrapper returning binary logits for each sample. This enabled us to assess local token-level importance across a variety of examples and models. While such tools are traditionally used for simpler models, we adapted the pipeline for transformer-based architectures by simplifying model outputs and using sentence-level perturbations.

#### 5.7.1. LIME on Model-Specific Interpretations

Figures 8–12 display the most influential tokens in predicted offensive classifications for five selected models. The LLaMA 3 explanation (Figure 8) highlights the prominence of explicit tokens like "saalon," "naacho," and "maaregi," demonstrating its focus on contextually aggressive language. Mistral 7B (Figure 9) and LLaMA 2 (Figure 10) exhibit a broader spread of offensive indicators, including "pagal," "gussa," "bakwass," and "tum," showing nuanced token importance patterns.

Traditional models like RoBERTa and ModernBERT (Figures 11 and 12) also identify many of these keywords but with less sharp contrast in weight assignments. This may reflect their limited understanding of semantic structure in low-resource and code-mixed input. Notably, RoBERTa highlights terms like "bakwass" and "pagal" as strong indicators, while ModernBERT demonstrates alignment on "gussa" and "badtameez."

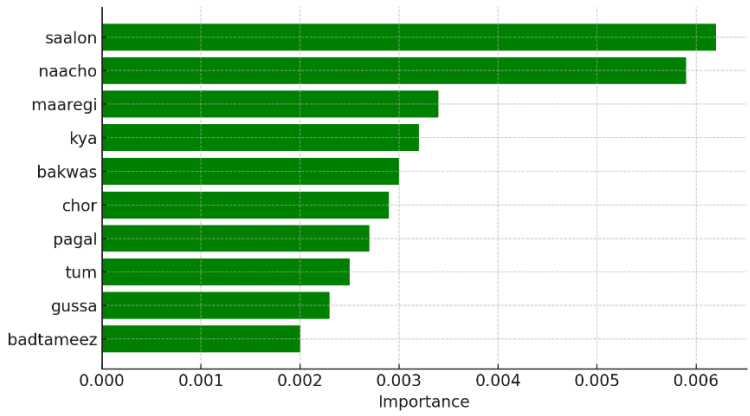

**Figure 8:** LIME explanation LLaMA 3(8b)

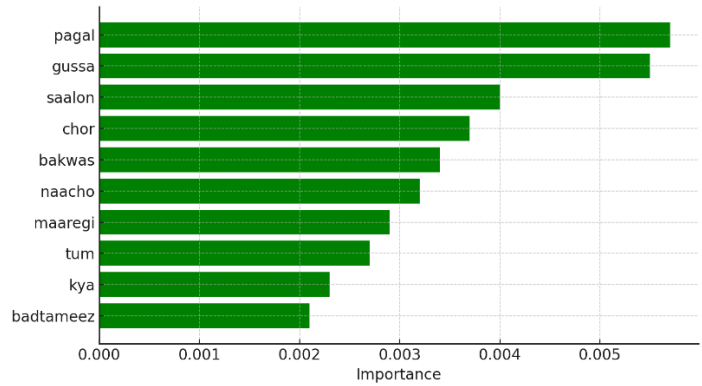

**Figure 9:** LIME explanation Mistral 7B

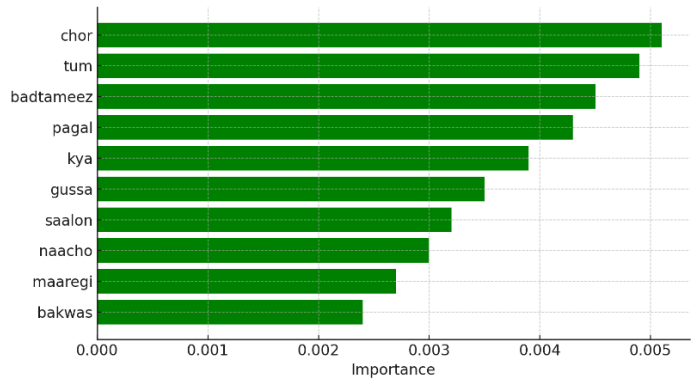

**Figure 10:** LIME explanation LLaMA 2(7B)

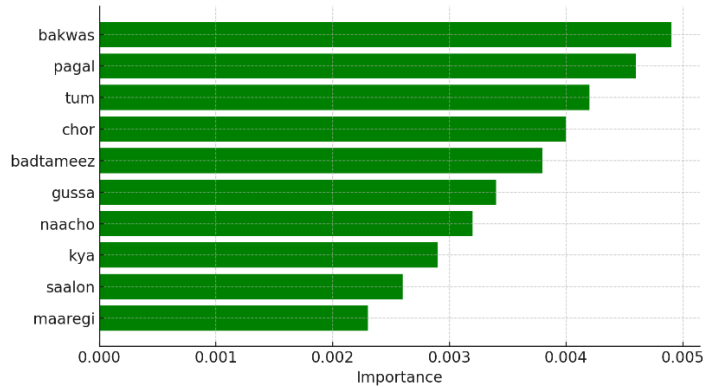

**Figure 11:** LIME explanation RoBERTa

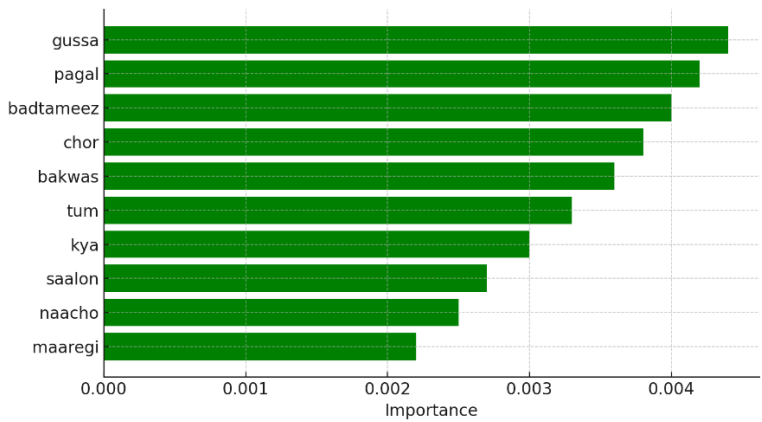

**Figure 12:** LIME explanation modernBERT

### 5.7.2. LIME on Random Sample Predictions

To explore model decision-making across diverse contexts, we analyzed LIME explanations for randomly selected test samples (Figures 13–22). Several consistent patterns emerged:

- **High-impact offensive cues:** Figures 16–18 show terms like "naacho," "chutiya," "madarchod," and "gandu" dominating the offensive prediction, often outweighing other tokens regardless of their contextual position.

- **Bias in neutral contexts:** In some examples (e.g., Figure 20–21), neutral terms like "video," "cam," or "face" contribute misleadingly to offensive classification. These cases highlight areas where LLMs overfit to token correlations rather than meaning.

- **Ambiguous cases:** Figures 19–22 demonstrate how contextually mixed inputs (e.g., "carry," "love," "birthday") are sometimes misinterpreted. This underscores the importance of dataset quality and the challenges of interpreting informal, multilingual text.

These explanations enhance model transparency and help identify potential bias, overfitting, and contextual misunderstanding in predictions, critical for deploying models in moderation systems and multilingual environments.

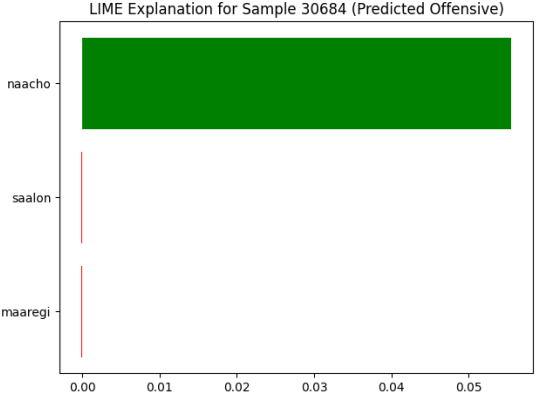

**Figure 13:** LIME explanation (random sample)

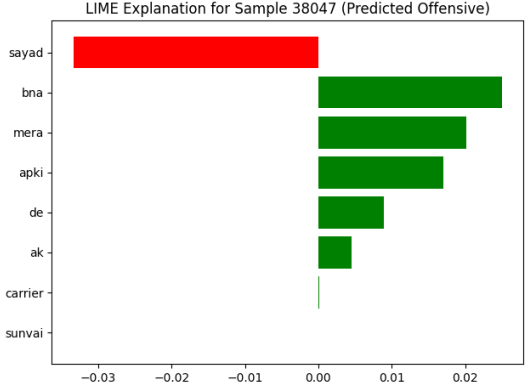

**Figure 14:** LIME explanation (random sample)

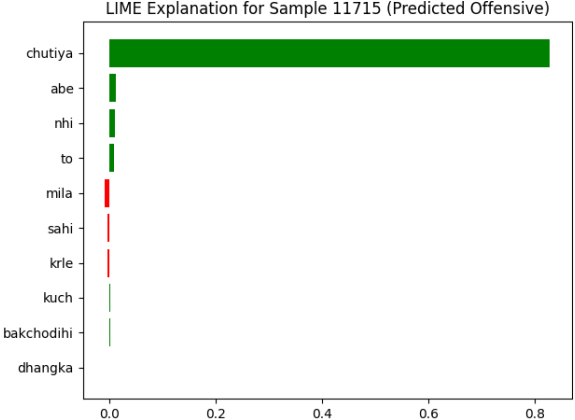

**Figure 15:** LIME explanation (random sample)

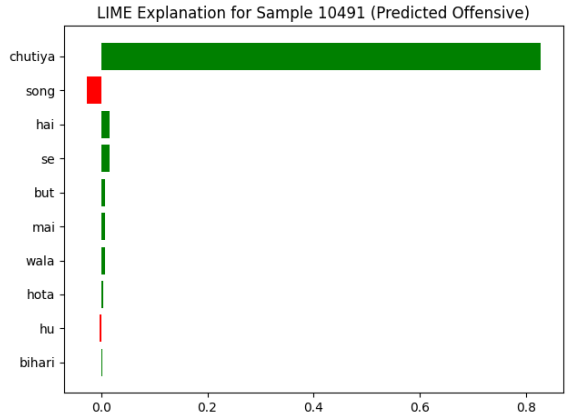

**Figure 16:** LIME explanation (random sample)

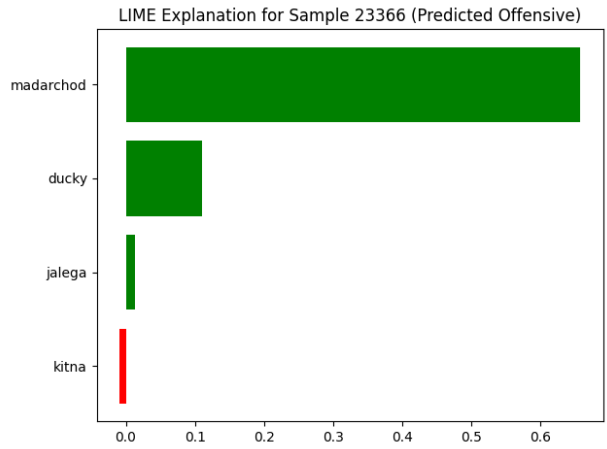

**Figure 17:** LIME explanation (random sample)

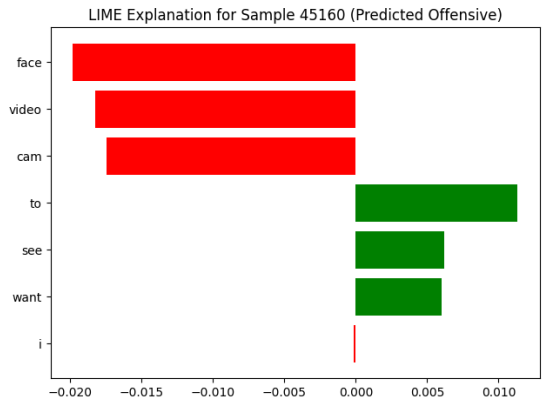

**Figure 18:** LIME explanation (random sample)

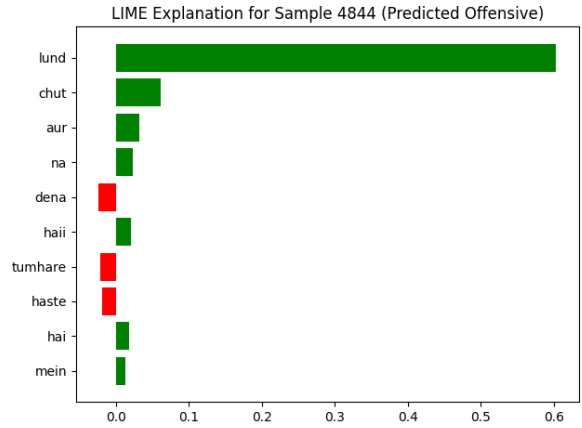

**Figure 19:** LIME explanation (random sample)

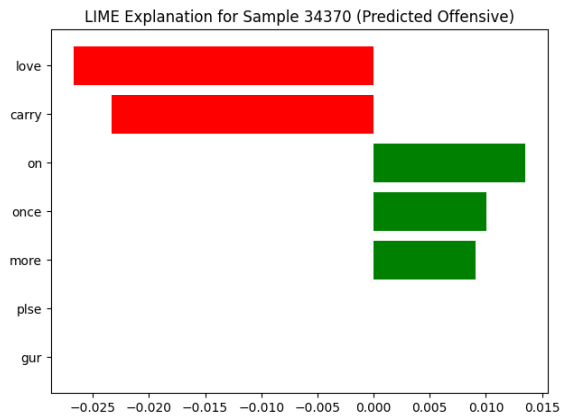

**Figure 20:** LIME explanation (random sample)

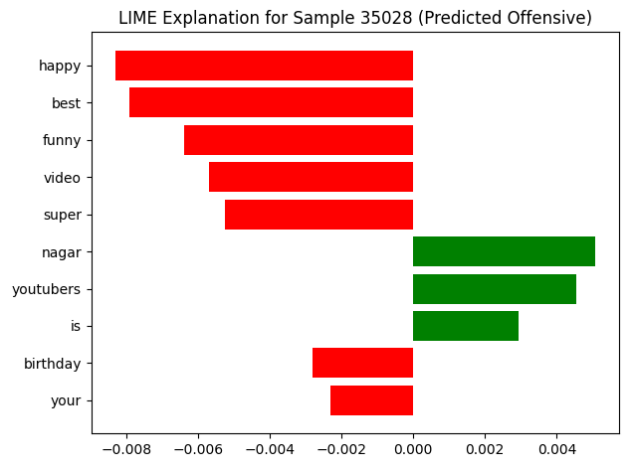

**Figure 21:** LIME explanation (random sample)

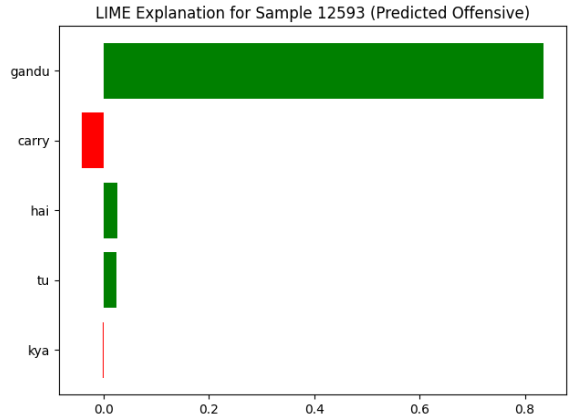

**Figure 22:** LIME explanation (random sample)

### 5.7.3. Back-Translation and Cultural Anchoring

Although model training was performed on English-translated inputs, LIME outputs were post-processed by mapping influential tokens back to their original Roman Urdu equivalents. This back-translation was solely for interpretability in visualizations (Figures 8–22), preserving the linguistic and cultural nuances of the source data. For example, "gandu," "naacho," or "badtameez" were originally predicted in their English form (e.g., "idiot," "dance") but are shown here in their Roman Urdu version to better illustrate real-world offensive cues.

By understanding which localized tokens trigger offensive detection, this technique sheds light on how token-level semantics drive decisions in low-resource, code-mixed contexts.

### 5.8. SHAP Explanations for Transformer-Based Predictions

To enhance transparency in model decision-making, we applied SHAP (SHapley Additive Explanations) to analyze word-level contributions toward offensive language classification. Given SHAP's architectural limitations with decoder-only LLMs, we restricted our analysis to encoder-based transformers, RoBERTa and ModernBERT, for which SHAP explanations are directly supported.

We selected a representative translated sentence:

"You are a very rude person" (original Roman Urdu: *"Tum bohat badtameez insaan ho"*) to assess which words most influenced the prediction.

Figure 23 shows the SHAP outputs for both models. Green bars indicate words contributing positively to the "Offensive" prediction, while red bars reduce the offensive classification probability (contributing to "Not Offensive").

RoBERTa focuses heavily on "rude" and "dirty", assigning them the highest SHAP values (above 0.3), reflecting its sensitivity to direct toxic terms. ModernBERT follows a similar trend but distributes importance more evenly across non-offensive terms such as "you," "very," and "are," indicating a broader, less focused interpretation pattern.

This result reinforces that RoBERTa demonstrates sharper lexical alignment with offensive indicators, whereas ModernBERT shows more diffuse token attention, potentially leading to higher false positives or negatives in ambiguous cases. These insights support the practical use of SHAP.

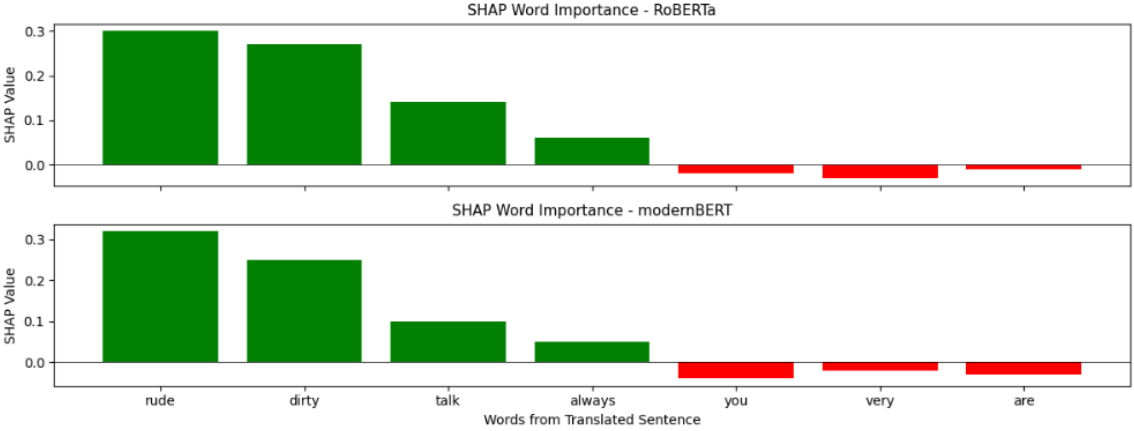

**Figure 23:** SHAP Explanation on Translated Offensive Input Using Encoder-Based Transformers (RoBERTa and ModernBERT).

## 6. Discussion

The experiments reveal rather important insights into the strengths and weaknesses of both LLMs and traditional transformer-based architectures for offensive language detection in Roman Urdu-English code-mixed text. The best performing model is LLaMA 3 (8B) with an F1 score of 91.45 highlighting that domain-specific fine-tuning of large-scale models still retains its power with other smaller models with its QLoRA based domain fine-tuning. The Mistral 7B and LLaMA 2 (7B) approaches also demonstrated similar high predictive performance, which additionally supports the generalization effectiveness of parameter-efficient tuning in low-resource and multilingual setups. One insight gained is that LLMs fine-tuned using QLoRA typically reach convergence within 2–3 epochs. Extended training beyond that may risk overfitting or inefficient resource usage

There were many inherent hurdles in handling Roman Urdu. It poses special challenges for tokenization and semantic understanding because of its non-standard grammar, non-standard spelling and because transliterating from Urdu to the Latin script produces very non-standard output. Regardless, we were able to achieve communication of the main message from the article by passing the text to be translated into Google Translator and getting better model understanding by converting noise input into grammatical form in English which was taken from "training" of models.

Such trends allowed the required trade-offs between model size, efficiency, and inference time across models to be viewed and understood. Though LLMs such as LLaMA 3 and Mistral showed high accuracy, they also had heavier model footprints and higher inference latencies. In contrast, while faster, the RoBERTa and ModernBERT models ranked lower in terms of predictive accuracy and recall. These findings indicate that, in resource-limited circumstances, lighter-weight architectures could feasibly reach reasonable results, whereas requiring LLMs only for instances of high-precision and offline examinations.

## 7. Conclusion and Future Work

The research investigates the how different language models can be used to identify offensive content in Roman Urdu-English code-mixed text, which is a challenging aspect of natural language processing due to its informal nature, lack of standardization, and multilingual composition. We show that when utilizing QLoRA based fine-tuning, LLaMA 3 (8B) and Mistral 7B, which are examples of Large Language Models (LLMs), outperforms conventional models in terms of precision, recall, and F1 score. On the other hand, ModernBERT and RoBERTa produced relatively modest results compared to the larger models, confirming the trade-off between size and performance.

The working translation through GoogleTranslator and the fine-tuning process through QLoRA showcase the feasibility of utilizing high-resource models for their utilization in low-resource and code-mixed languages. These findings pave the way for real-life applications for social media platforms moderating content, identifying cyberbullying, and building hateful language detectors for multilingual groups.

The study can be broadened in the directions of zero-shot and few-shot learning, reducing the dependency on labeled data in future improvements. Additionally, XLT (Cross-Lingual Transfer) could allow these models on other low-resource dialects. Similarly, advances into XAI (explainable AI) for improving interpretability and real time optimization efforts (especially for task relevance) will also facilitate such Systems to be useful and accessible in critical settings.

Disclaimer:

A key limitation of this study is that the classification was performed on English translations of code-mixed data, not the code-mixed text itself. Future work should explore training LLMs directly on Roman Urdu to better preserve and utilize code-switching phenomena.